\documentclass{icis}
\usepackage{authblk} 
\usepackage{graphicx}
\usepackage{xcolor}
\usepackage{amsmath}
\usepackage{tabularray} 
\title{Sentiment-Aware Extractive and Abstractive \\ Summarization for Unstructured Text Mining}
\researchtype{Regular Paper}
\shorttitle{Sentiment-Aware Summarization for Unstructured Text}
\conference{Workshop on Information Technologies and Systems, Nashville, USA 2025}

\usepackage[hidelinks]{hyperref}
\usepackage{tabularx}  
\usepackage{subfigure}
\usepackage{multirow}

\usepackage{dsfont}

\begin{document}

\maketitle


\bgroup
\begin{table}[!tbh]
\begin{tblr}{
colspec={Q[c,wd=0.45\textwidth] Q[c,wd=0.45\textwidth]},
width=\textwidth,
rowsep = 1pt,    
rows = {font=\fontsize{13pt}{15pt}\selectfont} 
}
\textbf{Junyi Liu}   &  \textbf{Stanley Kok}    \\
 Department of Information Systems and Analytics
 & Department of Information Systems and Analytics \\
National University of Singapore & National University of Singapore \\
 junyiliu@u.nus.edu & skok@comp.nus.edu.sg
    \end{tblr}
    \caption*{}
\end{table}
\vspace{-0.3in}
\egroup



  

\section{Introduction}


With the growing volume of unstructured textual data—such as social media posts, online reviews, forum discussions, and news articles—text mining has become increasingly crucial in Information Systems~\autocite{debortoli2016text}. It enables the automated extraction of implicit, previously unknown, and potentially valuable knowledge from vast datasets of unstructured data. Text mining techniques, including sentiment analysis, topic modeling, and text classification, equip organizations with the scalability and repeatability needed to process large datasets efficiently. Consequently, text mining has become an invaluable resource for businesses to understand consumer preferences, identify emerging trends, and respond to public sentiment in real-time.

Information systems (IS) research has increasingly leveraged text mining for organizational insight. For example, \textcite{qiao2023text} examined how incentivized reviews affect textual coherence; \textcite{buschken2016sentence} introduced sentence-level topic modeling; and \textcite{chen2024structural} incorporated sentiment signals into topic inference, highlighting how emotional undercurrents shape user opinions and inform business strategy. However, a significant challenge is handling the vast and fragmented unstructured data from platforms like Weibo, Instagram, and Reddit. The discussions on these platforms are wide-ranging, fragmented, and loosely structured, making it difficult for traditional text mining methods to efficiently extract useful information, especially in terms of capturing emotions and thematic contexts.

Text summarization offers a natural solution for handling vast and fragmented text by condensing lengthy content into concise summaries. It generally involves two approaches: extractive, which selects important sentences directly from the text, and abstractive, which generates new sentences that capture the essence of the original content. Existing summarization methods are optimized for structured texts like news articles but struggle with informal social media content. Social media often contains fragmented text and emotional expressions, making summarization more challenging. In Information Systems, understanding emotional cues is crucial for brand monitoring, customer interaction, and market analysis. Emotionally charged comments can reveal consumer feedback and potential issues, while ignoring these signals may lead to missed insights.

Although sentiment analysis has gained traction in IS for detecting emotional polarity and intensity ~\autocite{fang2014domain, oh2023you}, few studies have integrated sentiment modeling with summarization in the context of short, noisy, user posts. In practice, decision-makers need to understand not only what users discuss, but how they feel—whether satisfied, frustrated, or indifferent. Sentiment-enriched summaries can thus provide actionable insights, enabling timely interventions and strategic responses.

To address these challenges, we propose a sentiment-aware summarization framework for unstructured text mining. We extend both extractive summarization (e.g., graph-based TextRank) and abstractive summarization (e.g., Transformer-based UniLM) by integrating sentiment signals into their core mechanisms. 
This dual approach provides feasible solutions for improving both extractive and abstractive summarization methods, offering valuable contributions to text mining and Information Systems, particularly in enhancing the ability to capture emotional nuances and thematic relevance in user-generated content for decision support and strategic insights.

\section{Related Work}




\subsection{Text Mining in IS}

Text mining plays a crucial role in Information Systems (IS) research, especially due to its ability to analyze large volumes of unstructured textual data and extract valuable insights. Studies have investigated how text mining can help organizations understand consumer sentiment, inform business strategies, and improve predictive modeling. For example, research has explored how sentiment analysis of online reviews can influence purchasing behavior ~\autocite{yi2019leveraging}, how topic modeling on business pages can predict engagement patterns ~\autocite{yang2019understanding}, and how text classification methods applied to social media data can forecast trends more accurately than traditional approaches ~\autocite{song2019using}. Additionally, studies have examined how text mining techniques can track shifts in customer opinions~\autocite{goes2014popularity}, assess competition within digital content ecosystems ~\autocite{sabnis2015cable}, and measure the impact of external incentives on user-generated content quality ~\autocite{liu2021monetary}.

While these studies highlight the strategic potential of text mining, many focus on analyzing data volume, user behavior, or predictive accuracy. However, there has been less emphasis on developing automated methods for synthesizing short-form, fragmented, and sentiment-rich text—despite its increasing presence on platforms such as Reddit, Twitter, or Weibo. This gap limits the full potential of text mining in decision support systems (DSS), where decision-makers often need to interpret large datasets of user opinions quickly and accurately.

\subsection{Sentiment Analysis in IS}

Sentiment analysis has emerged as a powerful tool for understanding emotional cues in textual data. Within IS, prior work has addressed domain adaptation for sentiment classification~\autocite{fang2014domain} and examined how emotional tone influences content consumption and sharing behavior~\autocite{oh2023you}. For instance, negative sentiment has been shown to attract more attention on news websites, while positive sentiment increases shareability on social platforms.

Despite these advances, sentiment analysis is rarely integrated with summarization in IS research. Most approaches treat sentiment classification and summarization as separate tasks, overlooking their joint potential to distill both the content and emotional tone of UGC. This disconnect is particularly limiting in settings where the emotional valence of content—such as frustration, enthusiasm, or sarcasm—is key to interpreting user concerns or opportunities.

\section{Background}

\subsection{Text Summarization}

Text summarization aims to distill the essential content of a document into a concise version. Existing methods are broadly categorized into extractive and abstractive approaches. Extractive methods identify and select salient sentences from the input, ensuring grammaticality but often including redundancy. Abstractive methods generate new sentences that capture the core meaning, offering greater flexibility but requiring robust language modeling capabilities. Hybrid approaches aim to balance the strengths of both. 

\subsection{Textrank}

TextRank is a graph-based algorithm widely used for extractive summarization~\autocite{mihalcea2004textrank}. It treats the sentences of a document as nodes in a graph, where edges represent the similarity between sentences. The importance of each sentence is determined by its centrality within the graph.

Formally, given a document \( D = \{s_1, s_2, \dots, s_N\} \), a graph \( G = (V, E) \) is constructed, where \( V \) represents the sentences in \( D \) and \( E \) represents the edges connecting similar sentences. The weight of an edge between two sentences \( s_i \) and \( s_j \), denoted as \( w_{ij} \), is calculated as:
\[
w_{ij} = \frac{\text{sim}(s_i, s_j)}{\sum_{k \neq i} \text{sim}(s_i, s_k)},
\]
where \( \text{sim}(s_i, s_j) \) is a similarity function (e.g., cosine similarity or Jaccard index) that measures the relationship between sentences. The importance score \( p_i \) of a sentence \( s_i \) is computed iteratively:
\[
p_i = (1 - d) + d \cdot \sum_{j \in N(i)} \frac{w_{ij} \cdot p_j}{\sum_{k \in N(j)} w_{jk}},
\]
where \( d \) is a damping factor (typically set to 0.85), and \( N(i) \) represents the neighboring sentences of \( s_i \) in the graph. Sentences with the highest importance scores are selected for the summary.

\subsection{UniLM}



UniLM (Unified Language Model) is a pre-trained transformer model designed for both understanding and generation tasks~\autocite{dong2019unified}. In the context of summarization, UniLM generates abstractive summaries by conditioning on the input document. UniLM adopts a unified language modeling framework, in which different self-attention masks are used during training to support various language modeling objectives.

For abstractive summarization, UniLM is fine-tuned in an autoregressive sequence-to-sequence setting. Given an input document \( D \) and a summary sequence \( S = (s_1, s_2, \dots, s_{|S|}) \), the model is trained to maximize the conditional likelihood of each summary token given the document and previously generated tokens:
\[
\mathcal{L}
= - \sum_{t=1}^{|S|} \log P(s_t \mid s_1, s_2, \dots, s_{t-1}, D).
\]

At inference time, the summary is generated by solving
\[
S = \arg \max_S P(S \mid D),
\]
where \( P(S \mid D) \) denotes the conditional probability of the summary given the input document. This training objective enables the model to produce summaries that are coherent and relevant to the input text.

\subsection{Problem Formulation}

Our research aims to leverage emotion in text mining to achieve better text summarization. In this context, summarization not only focuses on content relevance but also incorporates user emotions to enhance the summary. Formally, given a document \( D = \{s_1, s_2, \dots, s_N\} \), summarization seeks a subset \( S \subseteq D \) that maximizes an objective function measuring content relevance, while also accounting for user emotion.


\section{Our Proposed Models}

We propose two methods for sentiment-aware text summarization: one for extractive summarization, ECPE-TextRank, which incorporates emotion-aware reasoning and topic relevance, and one for abstractive summarization, Senti-UniLM, which integrates sentiment-aware supervision into a Transformer-based model. Both methods aim to improve summarization by considering both factual content and emotional context. The framework can be seen as Figure 1.

\begin{figure}[h]
	\[
	\begin{array}{|c|}
	\hline \\ [-11pt]
	\includegraphics[scale = 0.45]{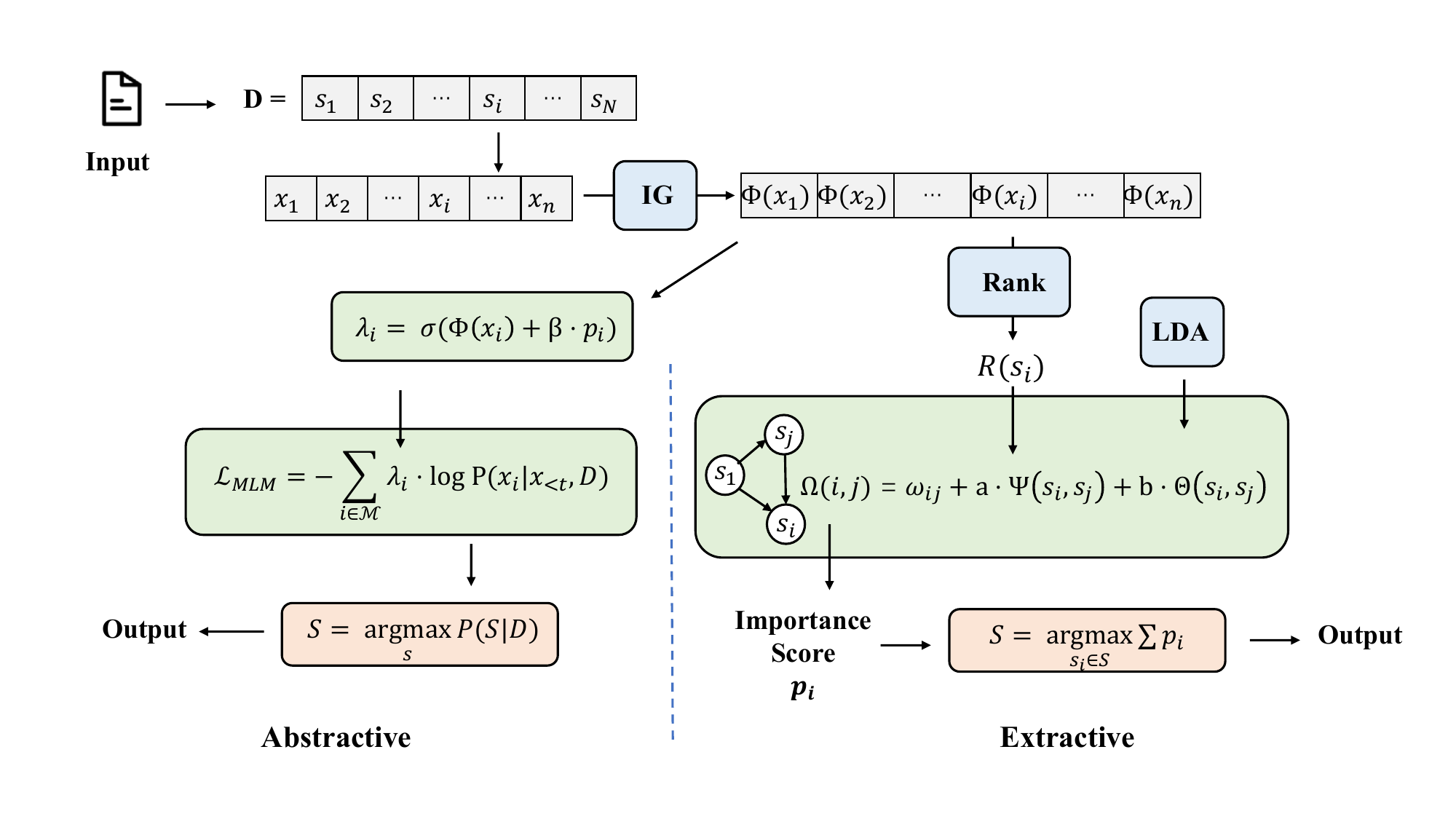} \\ [-4pt]
	\hline
	\addstackgap[7pt]{{\usefont{T1}{ptm}{b}{n}Figure 1.\hspace{0.09cm} Our Model Framework.}}\\
	\hline
	\end{array}
	\]
\end{figure}

\subsection{ECPE-TextRank}



In the original TextRank algorithm, sentence similarity is determined solely by lexical similarity $w_{ij}$, which limits its ability to capture emotional and thematic relevance. To address this limitation, we propose ECPE-TextRank, which replaces $w_{ij}$ with a composite similarity function $\Omega(s_i, s_j)$ integrating emotion-level, topic-level, and lexical-level similarities.

ECPE-TextRank incorporates Emotion-Cause Pair Extraction (ECPE) to identify emotional expressions and their underlying causes. Token-level sentiment importance is computed using the Integrated Gradients (IG) attribution method. For a tokenized sentence $S = \{x_1, x_2, \dots, x_n\}$, where each token is represented by its embedding, and a sentiment prediction model $F$, the attribution for token $x_i$ is defined as:
\[
\Phi_i(\mathbf{x}) = (x_i - x'_i) \cdot \int_{\alpha=0}^1 \frac{\partial F(\mathbf{x'} + \alpha (\mathbf{x} - \mathbf{x'}))}{\partial x_i} \, d\alpha,
\]
where $\mathbf{x'}$ denotes a baseline input (e.g., a zero embedding). The integral is numerically approximated using $m$ discrete steps $\alpha_j = \frac{j}{m}$:
\[
\Phi_i(\mathbf{x}) \approx (x_i - x'_i) \cdot \frac{1}{m} \sum_{j=1}^m \frac{\partial F(\mathbf{x'} + \alpha_j (\mathbf{x} - \mathbf{x'}))}{\partial x_i}.
\]

The resulting $\Phi(\mathbf{x})$ values are ranked, and the top-$k$ tokens form the sentiment cause set $R(S)$, representing a compact subset of salient words contributing to the sentiment prediction. Emotional similarity between two sentences $s_i$ and $s_j$ is measured as the cosine similarity between the embeddings of their cause sets:
$$
\Psi(s_i, s_j) = \cos\left( \mathbf{v}(R(s_i)), \mathbf{v}(R(s_j)) \right),
$$
where $\mathbf{v}(\cdot)$ denotes the mean embedding of tokens in the corresponding cause set.

To incorporate thematic relevance, topic similarity $\Theta(s_i, s_j)$ is computed via Latent Dirichlet Allocation (LDA)~\autocite{blei2003latent}. Each sentence is represented as a mixture of $K$ topics, with $p_k(s)$ denoting the probability of topic $k$ in sentence $s$:
$$
\Theta(s_i, s_j) = \sum_{k=1}^{K} p_k(s_i) \, p_k(s_j).
$$
The final similarity function combines lexical, emotional, and topic similarities:
$$
\Omega(s_i, s_j) =  w_{ij} + a \cdot \Psi(s_i, s_j) + b \cdot \Theta(s_i, s_j),
$$
where $a$ and $b$ are weights. Sentence importance scores $p_i$ are computed iteratively:
$$
p_i = (1 - d) + d \cdot \sum_{j \in N(i)} \frac{\Omega(s_i, s_j) \cdot p_j}{\sum_{k \in N(j)} w_{jk}},
$$
with damping factor $d$ (typically 0.85) and $N(i)$ as the neighbors of $s_i$. Sentences with the highest $p_i$ values are selected for the summary.

By replacing \(w_{ij}\) with \(\Omega(s_i, s_j)\), ECPE-TextRank incorporates emotional and thematic information into the ranking process. The final extractive summary \(S\) is obtained by selecting the top-\(L\) sentences with the highest importance scores \(p_i\):
\[
S = \arg\max_{S \subset D,\, |S| = L} \sum_{s_i \in S} p_i,
\]
where \(D\) is the set of all sentences in the document.

\subsection{Senti-UniLM}

UniLM uses Masked Language Modeling (MLM) to generate summaries, where the goal is to predict the masked word based on its surrounding context. 
While the objective above maximizes \(P(S \mid D)\) by treating all summary tokens equally during training, it may under-emphasize emotionally salient words. We therefore modify the training criterion by introducing token-specific sentiment weights into the sequence-to-sequence objective.

Specifically, each token \(x_i\) is assigned a weight \(\lambda_i\) that reflects its emotional importance. The weight is computed as a normalized combination of the token-level IG attribution score \(\Phi(x_i)\) and the sentence-level sentiment probability \(p(s)\) predicted by a sentiment prediction model, where \(x_i\) belongs to sentence \(s\):
\[
\lambda_i = \sigma\!\left(\Phi(x_i) + \beta\, p(s)\right),
\]
where \(\Phi(x_i)\) is the normalized IG score for token \(x_i\), \(p(s)\in(0,1)\) is the predicted sentiment probability for its containing sentence \(s\), \(\beta\) controls the influence of \(p(s)\), and \(\sigma(\cdot)\) maps the combined score into \((0,1)\).

The original sequence-to-sequence training objective is reformulated as a sentiment-weighted loss:
\[
\mathcal{L}_{\text{Senti}} = - \sum_{t=1}^{|S|} \lambda_t \log P(s_t \mid s_{<t}, D),
\]
which increases the training emphasis on emotionally salient tokens while preserving the autoregressive generation framework.

\section{Experiments}

\subsection{Dataset}

We evaluate our models on two benchmark datasets: Reddit-TIFU and DialogSum. The Reddit-TIFU dataset consists of 40,297 user-generated forum posts, with each entry including a title, user-written summary, and main content. 

DialogSum contains 13,460 multi-turn dialogues from daily scenarios like healthcare, education, and travel. All dialogues are preprocessed into BERT-compatible format, with speaker roles annotated~\autocite{chen2021dialogsum}.

\subsection{Evaluation}

We adopt the ROUGE metric (Recall-Oriented Understudy for Gisting Evaluation)~\autocite{lin2004rouge}, which is widely used in the summarization community due to its ability to evaluate the lexical overlap between generated summaries and human-written reference summaries. Specifically, we report ROUGE-1 (unigram), ROUGE-2 (bigram), and ROUGE-L (longest common subsequence) scores. The ROUGE-N recall score is formally defined as:
\[
\text{ROUGE-N} = \frac{\sum_{g \in G} \min\big(c_g^{(s)},\, c_g^{(r)}\big)}{\sum_{g \in G} c_g^{(r)}}
\]
where \( G \) denotes the set of all \( n \)-grams in the reference summary, \( c_g^{(s)} \) and \( c_g^{(r)} \) are the counts of \( g \) in the system-generated summary and reference summary, respectively. This metric reflects the proportion of reference content successfully recalled by the generated output.

ROUGE-L is computed based on the length of the longest common subsequence (LCS) between a system summary \( S \) and a reference summary \( R \), denoted by \( \mathrm{LCS}(S, R) \). The recall-oriented ROUGE-L is given by:
\[
\text{ROUGE-L} = \frac{\mathrm{LCS}(S, R)}{|R|}
\]
These formulations ensure a rigorous and interpretable comparison between generated and human-authored summaries, balancing token-level and sequence-level alignment.

\subsection{Our Model and Baselines}

To evaluate the performance of our proposed models, ECPE-TextRank and Senti-UniLM, we compare them against several well-established baseline methods. 

TextRank~\autocite{mihalcea2004textrank} is a classic graph-based extractive summarization method that ranks sentences via an iterative algorithm on a similarity graph.

UniLM~\autocite{dong2019unified}, a Transformer-based sequence-to-sequence model, leverages masked-token and sequence-level pretraining for effective abstractive summarization.

Pegasus~\autocite{zhang2020pegasus} uses Gap Sentences Generation, masking important sentences and reconstructing them to enhance abstractive capability.

BART~\autocite{lewis2021paq}, a denoising autoencoder, is pretrained by text corruption and reconstruction, enabling coherent abstractive summaries.

\subsection{Results}

We evaluate our models on two benchmark datasets, Reddit-TIFU and DialogSum, with the results summarized in Table 1.

\bgroup
\def\arraystretch{1.3}
\footnotesize
\begin{table}[ht]
\centering
\begin{tabular}{|c|>{\centering\arraybackslash}p{3.5cm}|>{\centering\arraybackslash}p{2.9cm}|>{\centering\arraybackslash}p{2.9cm}|>{\centering\arraybackslash}p{2.9cm}|}
\hline
\textbf{Dataset}    & \textbf{Model}           & \textbf{ROUGE-1} & \textbf{ROUGE-2} & \textbf{ROUGE-L} \\ \hline
\multirow{6}{*}{Reddit-TIFU} 
                    & TextRank                 & 0.5051          & 0.1232          & 0.4617          \\ \cline{2-5} 
                    & Bart-Finetune            & 0.2352          & 0.0619          & 0.1932          \\ \cline{2-5} 
                    & Pegasus                  & 0.1898          & 0.0589          & 0.1824          \\ \cline{2-5} 
                    & UniLM                    & 0.5449          & 0.1358          & 0.4456          \\ \cline{2-5} 
                    & ECPE-Textrank            & 0.5264          & 0.1347          & 0.4755          \\ \cline{2-5} 
                    & Senti-UniLM              & \textbf{0.6774} & \textbf{0.1824} & \textbf{0.5893} \\ \hline
\multirow{5}{*}{Dialogsum}    
                    & TextRank                 & 0.3191          & 0.1130          & 0.2985          \\ \cline{2-5} 
                    & Bart-Finetune            & 0.4285          & 0.1582          & 0.3765          \\ \cline{2-5} 
                    & Pegasus                  & 0.5317          & 0.3234          & 0.4172          \\ \cline{2-5} 
                    & UniLM                    & 0.5649          & 0.1862          & 0.4882          \\ \cline{2-5} 
                    & ECPE-Textrank            & 0.3530          & 0.1300          & 0.3316          \\ \cline{2-5} 
                    & Senti-UniLM              & \textbf{0.5893} & \textbf{0.1875} & \textbf{0.5160} \\ \hline
\multicolumn{5}{|c|}{\normalsize{\textbf{Table 1.} ROUGE evaluation results for various models on Reddit-TIFU and Dialogsum datasets.}} \rule{0pt}{3ex} \\ [4pt] \hline
\end{tabular}
\end{table}
\egroup

For the Reddit-TIFU dataset, both ECPE-TextRank and Senti-UniLM outperform their respective baselines (TextRank and UniLM), as well as other models such as Bart-Finetune and Pegasus, in all evaluation metrics. Specifically, Senti-UniLM achieves the highest ROUGE-1 score of 0.6774, significantly surpassing TextRank (0.5051), UniLM (0.5449), Bart-Finetune (0.2352), and Pegasus (0.1898). Similarly, both ROUGE-2 and ROUGE-L scores are higher for Senti-UniLM compared to all other models, demonstrating its superior performance in integrating Sentiment-aware features into the summarization process.

On the DialogSum dataset, Senti-UniLM also outperforms TextRank, UniLM, Bart-Finetune, and Pegasus, achieving a ROUGE-1 score of 0.5893, compared to TextRank (0.3191), UniLM (0.5649), Bart-Finetune (0.4285), and Pegasus (0.5317). These findings highlight that both ECPE-TextRank and Senti-UniLM provide significant improvements over traditional methods and approaches. 

\subsection{Ablation Experiment}


To evaluate the contribution of each component in the ECPE-TextRank model for sentiment-based text summarization, we conducted an ablation study involving two key modules: senti and topic. Three variants were tested—w/o senti, w/o topic, and w/o senti \& topic—by removing the specified module(s) while keeping others unchanged. 

\bgroup
\def\arraystretch{1.3}
\footnotesize
\begin{table}[ht]
    \centering
    \begin{tabular}{|>{\centering\arraybackslash}p{3.5cm}|>{\centering\arraybackslash}p{2.9cm}|>{\centering\arraybackslash}p{2.9cm}|>{\centering\arraybackslash}p{2.9cm}|}
        \hline
        \textbf{Model} & \textbf{ROUGE-1} & \textbf{ROUGE-2} & \textbf{ROUGE-L} \\ \hline
        Our Model      & \textbf{0.5264}  & \textbf{0.1347}  & \textbf{0.4755}  \\ \hline
        (w/o senti)    & 0.5160           & 0.1263           & 0.4662           \\ \hline
        (w/o topic)    & 0.5214           & 0.1328           & 0.4712           \\ \hline
        (w/o senti \& topic) & 0.5051    & 0.1232           & 0.4617           \\ \hline
        \multicolumn{4}{|c|}{\normalsize{\textbf{Table 2.} Ablation study results for ECPE-TextRank across ROUGE metrics.}} \rule{0pt}{3ex} \\ [4pt] \hline
    \end{tabular}
\end{table}
\egroup

As shown in Table 2, both w/o senti and w/o topic outperform w/o senti \& topic, but fall slightly short of the complete model, which achieves the highest scores across all ROUGE metrics (0.5264, 0.1347, and 0.4755). These results indicate that each module makes a distinct and complementary contribution, and their integration allows ECPE-TextRank to more effectively capture emotional nuances and thematic relevance in sentiment-aware summarization.


\subsection{Parameter Sensitivity Analysis}

In this subsection, we explore the sensitivity of the hyperparameters \(a\) and \(b\) in the ECPE-TextRank model. Specifically, we examine how varying \(a\) and \(b\) influences the performance of our model. In the experiments, we varied \(a\) and \(b\) from 0 to 1 with a step size of 0.05 and observed the resulting changes in performance.

\begin{figure}[h]
	\[
	\begin{array}{|c|}
	\hline 
	\includegraphics[scale = 0.35]{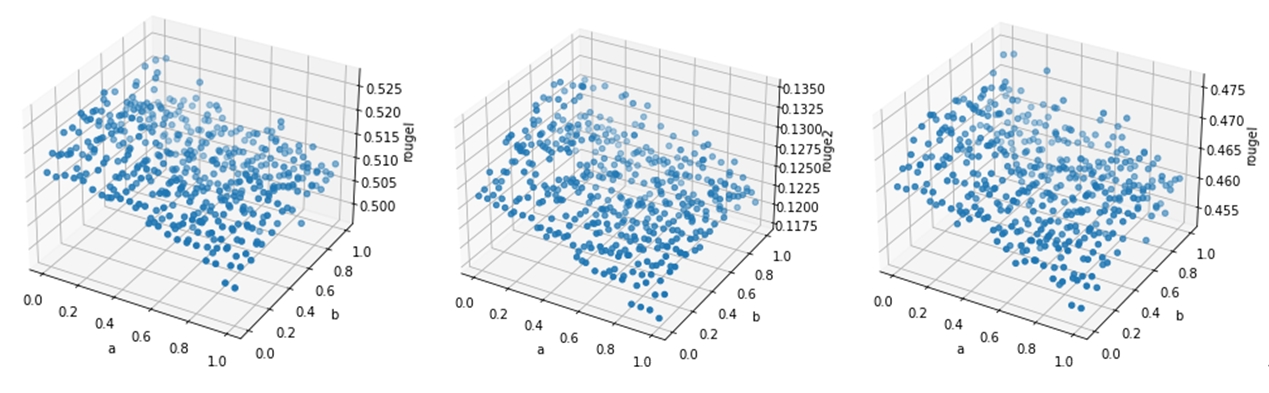} \\ [-2pt]
	\hline
	\addstackgap[7pt]{{\usefont{T1}{ptm}{b}{n}Figure 2.\hspace{0.09cm}  Illustrative of Parameter Sensitivity Analysis Experiment.}}\\
	\hline
	\end{array}
	\]
\end{figure}

The results, as illustrated in Figure 2, indicate that the model's performance is sensitive to both hyperparameters. The best performance is consistently achieved with $a = 0.1$, while the optimal value for $b$ varies depending on the evaluation metric: $b = 0.35$ for ROUGE-1, $b = 0.60$ for ROUGE-2, and $b = 0.15$ for ROUGE-L. Notably, increasing $a$ to larger values leads to a substantial decline in performance, suggesting that excessively high values for $a$ are detrimental to the model's effectiveness.

\section{Conclusion and Future Work}

In this work, we proposed a sentiment-aware summarization framework that combines extractive and abstractive methods to address the challenges of summarizing fragmented, emotion-rich user-generated content in Information Systems. By integrating sentiment and topic modules into ECPE-TextRank, our approach achieves superior performance over baselines and effectively captures both emotional nuances and thematic relevance. Ablation studies confirm the distinct and complementary contributions of each module. Future work will explore domain adaptation for diverse IS contexts, more sophisticated sentiment modeling (e.g., fine-grained emotion categories), and real-time deployment for large-scale social media monitoring, aiming to further enhance the practical value of sentiment-aware summarization in dynamic online environments.

\printbibliography

\end{document}